\documentclass{article}

\usepackage{times}
\usepackage{graphicx} 
\usepackage{subfigure} 
\usepackage{floatrow}
\usepackage{enumitem} 

\usepackage{natbib}

\usepackage{algorithm}
\usepackage{algorithmic}

\usepackage{hyperref}

\usepackage[accepted]{icml2017}

\usepackage{url}

\usepackage{siunitx}
\usepackage{amsmath}
\usepackage{graphicx}
\usepackage[export]{adjustbox}
\usepackage{graphbox} 

\usepackage[utf8]{inputenc} 
\usepackage[T1]{fontenc}    
\usepackage{booktabs}       
\usepackage{amsfonts}       
\usepackage{nicefrac}       
\usepackage{microtype}      

\usepackage{amssymb}
\usepackage{amsthm}
\usepackage{bbold}
\usepackage{graphicx}
\graphicspath{ {Figures/} }
\usepackage[export]{adjustbox}

\newcommand{\xx}{\mathbf{x}}
\newcommand{\hh}{\mathbf{h}}
\newcommand{\hht}{\mathbf{h}_t}
\newcommand{\xxt}{\mathbf{x}_t}

\newcommand{\Wx}{\mathbf{W_{\mb x_t}}}
\newcommand{\WW}{\mathbf{W}}

\newcommand{\bb}{\mathbf{b}}

\newcommand{\hhtm}{\mathbf{h}_{t-1}}

\newcommand{\pp}[1]{\left( #1 \right)}
\newcommand{\mb}{\mathbf}

\newcommand{\mkappa}{\boldsymbol{ \kappa}}

\newcommand{\norm}[1]{\left|\left| #1 \right|\right|}
\newcommand{\expect}[2]{\mathbb{E}_{#1}\left[ #2 \right]}

\newcommand{\bbx}{\mathbf{b_{\mb x_t}}}

\newcommand{\david}[1]{\textcolor{straightblingin}{[David: #1]}}
\newcommand{\jascha}[1]{\textcolor{darkgreen}{[Jascha: #1]}}
\newcommand{\jakob}[1]{\textcolor{red}{Jakob: \uppercase{#1}}}
\newcommand{\justin}[1]{\textcolor{blue}{Justin: #1}}
\newcommand{\jan}[1]{\textcolor{orange}{Jan: #1}}

\renewcommand{\david}[1]{ }
\renewcommand{\jascha}[1]{ }
\renewcommand{\jakob}[1]{ }
\renewcommand{\justin}[1]{}
\renewcommand{\jan}[1]{ }

\icmltitlerunning{Interpretable RNNs with Input Switched Affine Networks}

\begin{document} 

\twocolumn[
\icmltitle{Input Switched Affine Networks: An RNN Architecture Designed for Interpretability}



\icmlsetsymbol{equal}{*}

\begin{icmlauthorlist}
\icmlauthor{Jakob N. Foerster}{equal,gintern}
\icmlauthor{Justin Gilmer}{equal,gburp}
\icmlauthor{Jascha Sohl-Dickstein}{goob}
\icmlauthor{Jan Chorowski}{gfac}
\icmlauthor{David Sussillo}{goob}
\end{icmlauthorlist}

\icmlaffiliation{goob}{Google Brain, Mountain View, CA, USA}
\icmlaffiliation{gburp}{Work done as a member of the Google Brain Residency program (\url{g.co/brainresidency})}
\icmlaffiliation{gintern}{This work was performed as an intern at Google Brain}
\icmlaffiliation{gfac}{Work performed when author was a visiting faculty at Google Brain}

\icmlcorrespondingauthor{Jakob N. Foerster}{jakob.foerster@cs.ox.ac.uk}
\icmlcorrespondingauthor{David Sussillo}{sussillo@google.com}

\icmlkeywords{interpretable deep learning, recurrent neural networks, switched affine system}

\vskip 0.3in
]



\printAffiliationsAndNotice{\icmlEqualContribution} 

\begin{abstract}
There exist many problem domains where the interpretability of neural network models is essential for deployment. 
Here we introduce a recurrent architecture composed of input-switched affine transformations -- in other words an RNN without any explicit nonlinearities, but with input-dependent recurrent weights.  
This simple form allows the RNN to be analyzed via straightforward linear methods: we can {\em exactly} characterize the linear contribution of each input to the model predictions; 
we can use a change-of-basis to disentangle input, output, and computational hidden unit subspaces; 
we can fully reverse-engineer the architecture's solution to a simple task. 
Despite this ease of interpretation, the input switched affine network achieves reasonable performance on a text modeling tasks, and allows greater computational efficiency than networks with standard nonlinearities.

\end{abstract}
\section{Introduction}

\subsection{The importance of interpretable machine learning}

As neural networks move into applications where the outcomes of human lives depend on their decisions, it is increasingly crucial that we are able to interpret the decisions they make.  Indeed, the European Union is considering legislation with a clause that asserts that individuals have 'rights to explanation', i.e. individuals should be able to understand how algorithms make decisions about them \citep{eureg_2016_679}.  Example problem domains requiring interpretable ML include self-driving cars \cite{bojarski2016end}, air traffic control \cite{katz2017reluplex}, power grid control \cite{siano2012real}, hiring and promotion decisions while preventing bias \cite{scarborough2006neural}, automated sentencing decisions in US courts \cite{tashea2017,berk2017fairness}, and medical diagnosis \cite{gulshan2016development}. For many of these applications, practitioners will not adopt ML models without fully understanding what drives their predictions, including understanding when and how these models fail \cite{ching2017opportunities,deo2015machine}.  

\subsection{Post hoc analysis}

One approach to interpreting neural networks is to train the network as normal, and then apply analysis techniques after training. Often this approach yields systems that perform extremely well, but where interpretability is challenging.  For example, \citet{sussillo2013opening} used linearization and nonlinear dynamical systems theory to understand RNNs solving a set of simple but varied tasks. \citet{karpathy2015visualizing} analyzed an LSTM \cite{hochreiter1997long} trained on a character-based language modeling task. They were able to break down LSTM language model errors into classes, such as e.g., ``rare word'' errors. Concurrently with our submission, \citet{murdoch2017automatic} decomposed LSTM outputs using telescoping sums of statistics computed from memory cells at different RNN steps. The decomposition is exact, but not unique and the authors justify it by demonstrating good performance of decision rules formed using the computed cell statistics. 


The community is also interested in post hoc interpretation of feed-forward networks. Examples include
the use of linear probes in \citet{alain2016understanding}, and a variety of techniques (most driven by back-propagation) to assign credit for activations to specific inputs or input patterns in feed-forward networks \cite{zeiler2010deconvolutional,le2012building,mordvintsev2015inceptionism}.

\subsection{Building interpretability into the architecture}

A second approach is to build a neural network where interpretability is an explicit design constraint.  In this approach, a typical outcome is a system that can be better understood, but at the cost of reduced performance.  Model classes whose decisions are naturally interpretable include logistic regression \cite{freedman2009statistical}, decision trees \cite{quinlan1987simplifying}, and support vector machines with simple (e.g. linear) kernels \cite{andrew2013introduction}. 

In this work we follow this second approach and build interpretability into our network model, while maintaining good, though not always state-of-the-art, performance for the tasks we study. 
We focus on the commonly studied task of character based language modeling. We develop and analyze a model trained on a one-step-ahead prediction task of the Text8 dataset, which is 10 million characters of Wikipedia text \cite{text8}, on the Billion Word Benchmark \cite{2013arXiv1312.3005C}, and finally on a toy multiple parentheses counting task which we fully reverse engineer.

\subsection{Switched affine systems}

The model we introduce is an Input Switched Affine Network (ISAN), where the input determines the switching behavior by selecting a transition matrix and bias as a function of that input, and there is no nonlinearity. Linear time-varying systems are standard material in undergraduate electrical engineering text books, and are closely related to our technique.

Although the ISAN is deterministic, probabilistic versions of switching linear models with discrete latent variables have a history in the context of probabilistic graphical models.  A recent example is the switched linear dynamical system in \cite{linderman2016recurrent}.  Focusing on language modeling, \citep{belanger2015linear} defined a probabilistic linear dynamical system (LDS) as a generative language model for creating context-dependent token embeddings and then used steady-state Kalman filtering for inference over token sequences.  They used singular value decomposition and discovered that the right and left singular vectors were semantically and syntactically related.  A critical difference between the ISAN and the LDS is that the ISAN weight matrices are input token dependent (while the biases of both models are input dependent).

Multiplicative neural networks (MRNNs) were proposed precisely for character based language modeling in \citep{sutskever2011generating,martens2011learning}. 
The MRNN architecture is similar to our own, in that the dynamics matrix switches as a function of the input character. 
However, the MRNN relied on a $\tanh$ nonlinearity, while the ISAN is explicitly linear. It is this property of our model which makes it both amenable to analysis, and computationally efficient. 

The Observable Operator Model (OOM) \citep{jaeger2000observable} is similar to the ISAN in that the OOM updates a latent state using a separate transition matrix for each input symbol and performs probabilistic sequence modeling. Unlike the ISAN, the OOM requires that a linear projection of the hidden state corresponds to a normalized sequence probability. 
This imposes strong constraints on both the model parameters and the model dynamics, and restricts the choice of training algorithms. In contrast, the ISAN applies an affine readout to the hidden state to obtain logits, which are then pushed through the softmax function to obtain probabilities. Therefore no constraints need to be imposed on the ISAN's parameters and training is easy using backprop. Lastly, the ISAN is formulated as an affine, rather than linear model. While this doesn't change the class of processes that can be modeled, it stabilizes training and greatly enhances interpretability, facilitating the analysis in Section \ref{sec:kappa_decomposition}. 

\subsection{Paper structure}

In what follows, we define the ISAN architecture, demonstrate its performance on the one-step-ahead prediction task, and then analyze the model in a multitude of ways, most of which would be currently difficult or impossible to accomplish with modern nonlinear recurrent architectures.

\section{Methods}

\subsection{Model definition}

In what follows $\mb W_{\mb x}$ and $\mb b_{\mb x}$ respectively denote a transition matrix and a bias vector for a specific input $\xx$, the symbol $\xxt$ is the 
input at time $t$,  and $\hht$ is the hidden state at time $t$.  Our ISAN model is defined as 
\begin{align}
    \hht = \Wx \; \hhtm + \bbx.
\label{eq:ISAN_transition}
\end{align}  
The network also learns an initial hidden state $\hh_0$. We emphasize the intentional absence of any nonlinear activation function.
\subsection{Character level language modeling with ISAN}

We trained RNNs on the Text8 Wikipedia dataset and the billion word benchmark (BWB), for one-step-ahead character prediction. The Text8 dataset consists only of the 27 characters `a'-`z' and `\_' (space).  The BWB dataset consist of Unicode text and was modelled as a sequence of bytes (256 discrete tokens) that formed the UTF8-encoded data.
 Given a character sequence of $\mb x_1, ..., \mb x_t$, the RNNs are trained to minimize the cross-entropy between the true next character, 
 and the output prediction.  We map from the hidden state, $\hht$, into a logit space via an affine map. 
 The probabilities are computed as
\begin{align}
    p\pp{\mb x_{t+1}} &= \mbox{softmax}\pp{\mb l_t} \\
    \mb l_t &=     \WW_{ro}\;\hht + \bb_{ro},
\end{align}
where $\WW_{ro}$ and $\bb_{ro}$ are the readout weights and biases, and $\mb l_t$ is the logit vector.  
For the Text8 dataset, we split the data into 90\%, 5\%, and 5\% for train, validation, and test respectively, in line with \citep{mikolov2012subword}.
The network was trained with the same hyperparameter tuning infrastructure as in \citep{collins2016capacity}.  For the BWB dataset, we used data splits and evaluation setup identical to \citep{jozefowicz2016exploring}. Due to long experiment running times, we manually tuned the hyperparameters.

\section{Results and analysis}

\subsection{ISAN performance on Text8 prediction}
The results on Text8 are shown in Table \ref{table text8}.  For the largest parameter count, the ISAN matches almost exactly the performance of all other nonlinear models with the same number of maximum parameters: RNN, IRNN, GRU, LSTM. However, we note that for small numbers of parameters the ISAN performs considerably worse than other architectures. All analyses use ISAN trained with 1.28e6 maximum parameters (1.58 bpc cross entropy).  Samples of generated text from this model are relatively coherent. We show two examples, after priming with "annual reve", at inverse temperature of 1.5, and 2.0, respectively:
\begin{itemize}[topsep=0pt,itemsep=-1ex,partopsep=1ex,parsep=1ex]
    \item \textit{``annual revenue and producer of the telecommunications and former communist action and saving its new state house of replicas and many practical persons''}
    \item \textit{``annual revenue seven five three million one nine nine eight the rest of the country in the united states and south africa new''}.
\end{itemize}

\begin{table}[!t]
\centering
\caption{%
The ISAN has similar performance to other RNN architectures on the Text8 dataset. 
Performance of RNN architectures on Text8 one-step-ahead prediction, measured as cross-entropy loss on a held-out test set, in bits per character. The loss is shown as a function of the maximum number of parameters a model is allowed.  The values reported for all other architectures are taken from \citep{collins2016capacity}. }
\begin{tabular}[b]{|c | c c c|} 
 \hline
 Parameter count & 8e4 & 3.2e5 & 1.28e6 \\ [0.75ex]
 \hline\hline
 RNN       & 1.88 & 1.69 & 1.59 \\ 
 IRNN      & 1.89 & 1.71 & 1.58 \\
 GRU       & 1.83 & 1.66 & 1.59 \\
 LSTM      & 1.85 & 1.68 & 1.59 \\ 
 ISAN  & 1.92 & 1.71 & 1.58\\ [1ex] 
 \hline
\end{tabular} \label{table text8}
\end{table}

\begin{figure}[!t]
\includegraphics[clip,width=0.65\linewidth]{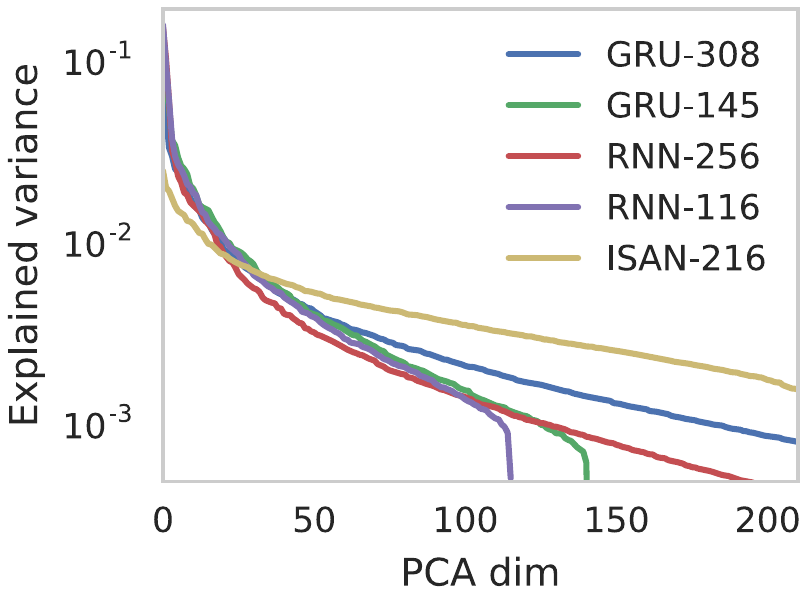} 
\caption{%
The ISAN makes fuller and more uniform use of its latent space than vanilla RNNs or GRUs. Figure shows explained variance ratio of the first 210 most significant PCA dimensions of the hidden states across several architectures for the Text8 dataset. The legend provides the number of latent units for each architecture.}
\label{fig rand}
\end{figure}

As a preliminary, comparative analysis, we performed PCA on the state sequence over a large set of sequences for the vanilla RNN, GRU of varying sizes, and ISAN.  This is shown in Figure \ref{fig rand}.  The eigenvalue spectra, in log of variance explained, was significantly flatter for the ISAN than the other architectures. 

We compared the ISAN performance to a fully linear RNN without input switched dynamics. This achieves a cross-entropy of ~3.1 bits / char, independent of network size. This perplexity is only slightly better than that of a Naive Bayes model on the task, at 3.3 bits / char. The output probability of the fully linear network is a product of contributions from each previous character, as in Naive Bayes. Those factorial contributions are learned however, giving the non-switched affine network a slight advantage. We also trained a fully linear network with a nonlinear readout. This achieves ~2.15 bits / char, independent of network size. Both of these comparisons illustrate the importance of the input switched dynamics for achieving good results.

Lastly we also test to what extent the ISAN can deal with large dictionaries by running it on a byte-pair encoding of the text8 task, where the input dictionary consists of the $27^2$ different possible character combinations. We find that in this setup the LSTM consistently outperforms the ISAN for the same number of parameters. At $1.3m$ parameters the LSTM achieves a cross entropy of 3.4 bits / char-pair, while ISAN achieves 3.55. One explanation for this finding is that the matrices in ISAN are 27 times smaller than the matrices of the LSTMs. For very large numbers of parameters the performance of any architecture saturates in the number of parameters, at which point the ISAN can `catch-up' with more parameter efficient architectures like LSTMs.

\begin{figure*}
\centering
\includegraphics[trim={0 0cm 0 0 0},clip,width=1\linewidth]{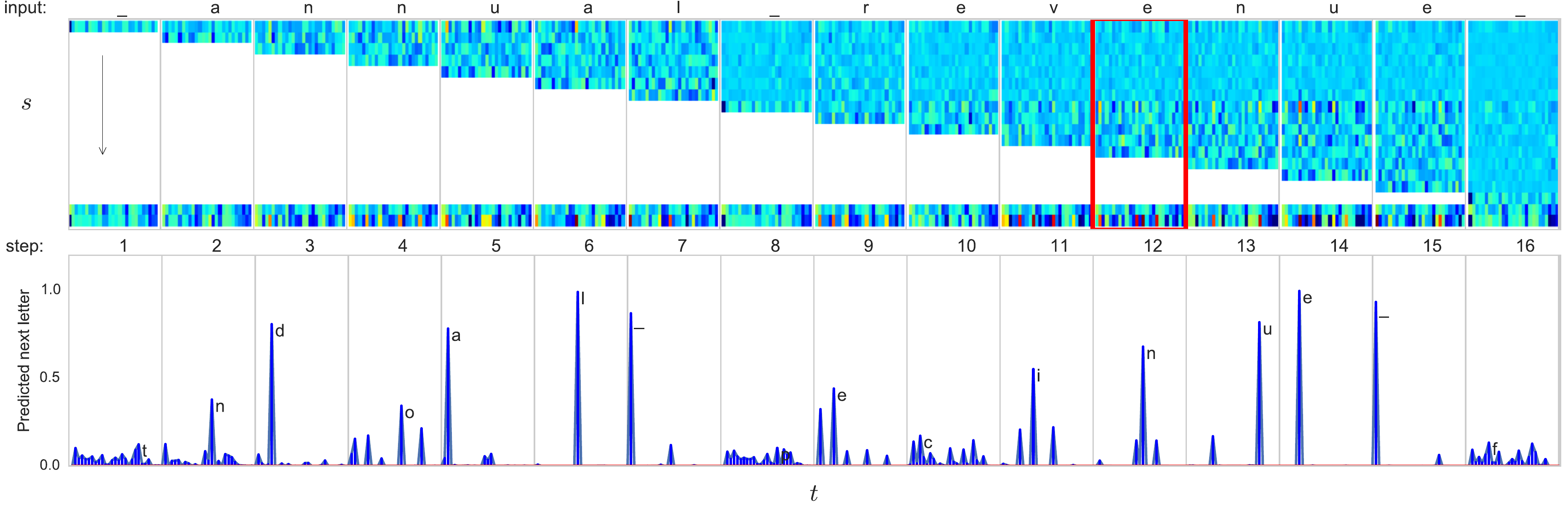}
\caption{Using the linearity of the hidden state dynamics, predictions at step $t$ can be broken out into contributions, $\mkappa^t_s$, from previous steps. Accordingly, each row of the top panel corresponds to the propagated contribution ($\mkappa^t_s$) of the input character at time $s$, to the prediction at time $t$ (summed to create the logit at time $t$). The penultimate row contains the output bias vector replicated at every time step. The last row contains the logits of the predicted next character, which is the sum of all rows above.  The bottom panel contains the corresponding softmax probabilities at each time $t$ for all characters (time is separated by gray lines). Labeled is the character with the maximum predicted probability. 
The time step boxed in red is examined in more detail in Figure \ref{fig:zoom_in}.
}
\label{fig:opening_the_black_box}
\end{figure*}

\subsection{ISAN performance on Billion Word Benchmark prediction}

We trained ISAN and LSTM models on the BWB dataset. All networks were trained using asynchronous gradient descent using the Adagrad learning rule. Our best LSTM model reached 1.1 bits per character, which matches published results \cite{hwang16character}. The LSTM model had one layer of 8192 LSTM units whose outputs were projected onto 1024 dimensions (44e6 parameters). Our best ISAN models reached 1.4 bits per character and used 512 hidden units, a reduced set of most common 70 input tokens and 256 output tokens (18e6 parameters). Increasing ISAN's hidden layer size to 768 units (41e6 parameters) yielded a perplexity improvement to 1.36 bits/char. Investigation of generated samples shows that the ISAN learned the distinction between lower- and upper-cased letters and is able to generate text which is coherent over short segments. To demonstrate sample variability we show continuations of the prompt "The [Pp]ol" generated using the ISAN:
\begin{itemize}[topsep=0pt,itemsep=-1ex,partopsep=1ex,parsep=1ex]
    \item \emph{The Pol|ish pilgrims are as angry over the holiday trip}
    \item \emph{The Pol|ice Department subsequently slipped toward}
    \item \emph{The Pol|ice Federation has sought Helix also investors}
    \item \emph{The Pol|itico is in a tight crowd ever to moderated the}
    \item \emph{The pol|itical scientist in the Red Shirt Romance cannot}
    \item \emph{The pol|icy for all Balanchine had formed when it set a}
    \item \emph{The pol|l conducted when a suspected among Hispanic}
    \item \emph{The pol|itical frenzy sparked primary care programs}
\end{itemize}

\subsection{Decomposition of current predictions based on previous time steps} \label{sec:kappa_decomposition}
Analysis in this paper is carried out on the best-performing Text8 ISAN model, which has $1,271,619$ parameters, corresponding to $216$ hidden units, and 27 dynamics matrices $\mb W_{\mb x}$ and biases $\mb b_{\mb x}$.  

With ISAN we can analyze which factors were important in the past for determining the current character prediction.  Taking advantage of the linearity of the hidden state dynamics for any sequence of inputs, we decompose the current latent state $\mb h_t$ into contributions originating from different time points $s$ in the history of the input:
\begin{align}
\mb h_t &= \sum_{s=0}^{t} \left( \prod_{s'=s+1}^{t} \mb W_{\mb x_{s'}}\right) \mb b_{\mb x_{s}}
,
\end{align}
where the empty product when $s+1 > t$ is $1$ by convention, and $\mb b_{\mb x_0} = \mb h_0$ is the learned initial hidden state. 

Using this decomposition and the fact that matrix multiplication is a linear transformation we can also write the unnormalized logit-vector, $\mb l_t$, as a sum of terms linear in the biases,
\begin{align}
\mb l_t &= \mb b_{ro} + 
    \sum_{s=0}^{t}
    \mkappa_s^t \\
\mkappa_s^t &= \mb W_{ro} \left( \prod_{s'=s+1}^{t} \mb W_{\mb x_{s'}}\right) \mb b_{\mb x_{s}}, \label{eq:kappa}
\end{align}
where $\mkappa_s^t$ is the contribution from time step $s$ to the logits at time step $t$, and $\mb \mkappa_t^t = \mb b_{\mb x_t}$.
For notational convenience we will sometimes replace the subscript $s$ with the corresponding input character $\mb x_s$ at step $s$ when referring to $\mkappa_s^t$.  For example, $\mkappa_{\text{`q'}}^t$ refers to the contribution from the character `q' in a string. 
Similarly, when discussing the summed contributions from a word or substring we will sometimes write $\mkappa_{word}^t$ to mean the summed contributions of all the $\mkappa_s^t$ from that source word. For example, $\sum_{s\in\text{word}}\mkappa_s^t$ -- $\mkappa_{\text{`the'}}^t$ refers to the total contribution from the word `the' to the logit.

While in standard RNNs the nonlinearity causes interdependence of the bias terms across time steps, in the ISAN the bias terms contribute to the state as independent linear terms that are propagated and transformed through time.
We emphasize that $\mkappa_{s}^t$ includes the multiplicative contributions from the $\mb W_{\mb x_{s'}}$ for $s < s' \leq t$. It is however independent of prior inputs, $\mb x_{s'}$ for $s' < s$. This is the main difference between the analysis we can carry out with the ISAN compared to a nonlinear RNN. In a general recurrent network the contribution of a specific character sequence will depend on the hidden state at the start of the sequence. Due to the linearity of the dynamics, this dependency does not exist in the ISAN.

In Figure~\ref{fig:opening_the_black_box} we show an example of how this decomposition allows us to understand why a particular prediction is made at a given point in time, and how previous characters influence the decoding. For example, the sequence `\_annual\_revenue\_' is processed by the ISAN: Starting with an all-zero hidden state, we use equation (\ref{eq:kappa}) to accumulate a sequence of $\mkappa^t_{`\_'}, \mkappa^t_{`a'}, \mkappa^t_{`n'},\mkappa^t_{`n'},...$. We then used these values to understand the prediction of the network at some time $t$, by simple addition across the $s$ index. 

\begin{figure}[t]
\centering
\includegraphics[trim={0 0 0 0},clip,width=1\linewidth]{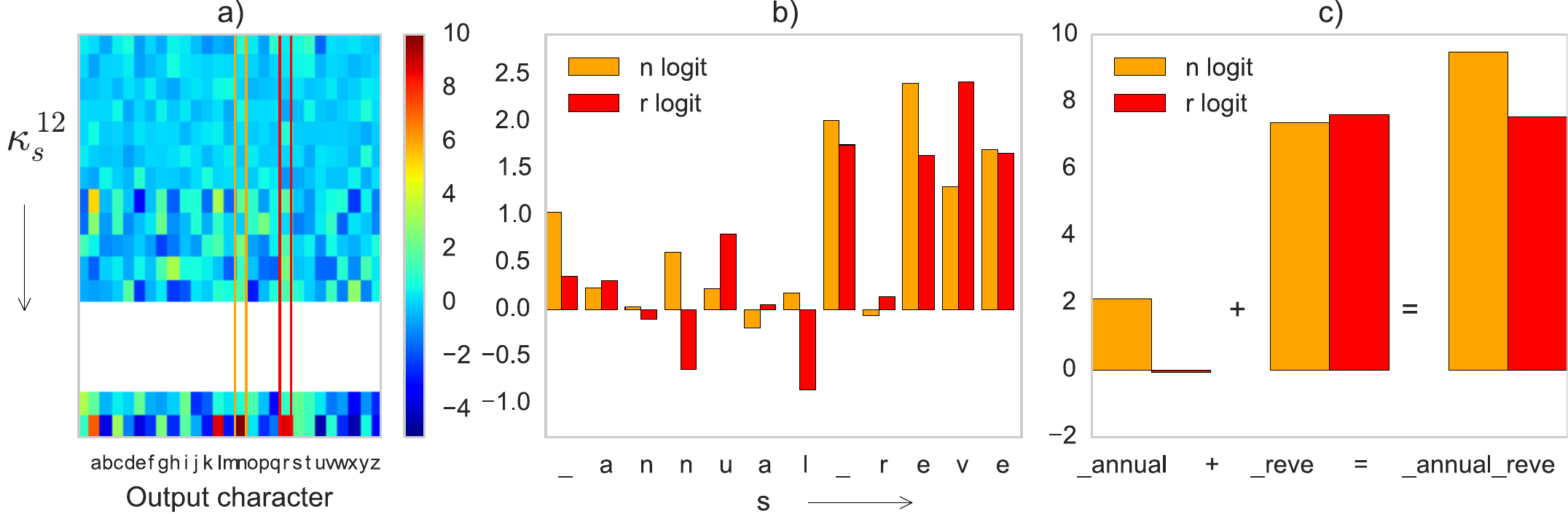}
\caption{Detailed view of the prediction stack for the final `n' in `\_annual\_reve\underline{n}ue'. In {\em a)} all $\mkappa^{t}_s$ are shown, in {\em b)} only the contributions to the `n' logit and `r' logits are shown, in orange and red respectively, from each earlier character in the string. This corresponds to a zoom in view of the columns highlighted in orange and red in a). In {\em c)} we show how the sum of the contributions from the string `\_annual', $\mkappa^t_\text{`\_annual'}$, pushes the prediction at `\_annual\_reve' from `r' to `n'. Without this contribution the model decodes based only on $\mkappa^t_\text{`\_reve'}$, leading to a MAP prediction of `reverse'. With the contribution from $\mkappa^t_\text{`\_annual'}$ it instead predicts `revenue'.  The contribution of $\mkappa^t_\text{`\_annual'}$ to the `n' and `r' logits is linear and exact.}
\label{fig:zoom_in}
\end{figure}

We provide a detailed view of how past characters contribute to the logits predicting the next character in Figure~\ref{fig:zoom_in}. There are two competing options for the next letter in the word stem `reve': either `reve\textbf{n}ue' or `reve\textbf{r}se'.  We show that without the contributions from `\_annual' the most likely decoding of the character after the second `e' is `r' (to form `reverse'), while the contributions from `\_annual' tip the balance in favor of `n', decoding to `revenue'.

\begin{figure}[t]
\centering
\includegraphics[width=0.95\linewidth]{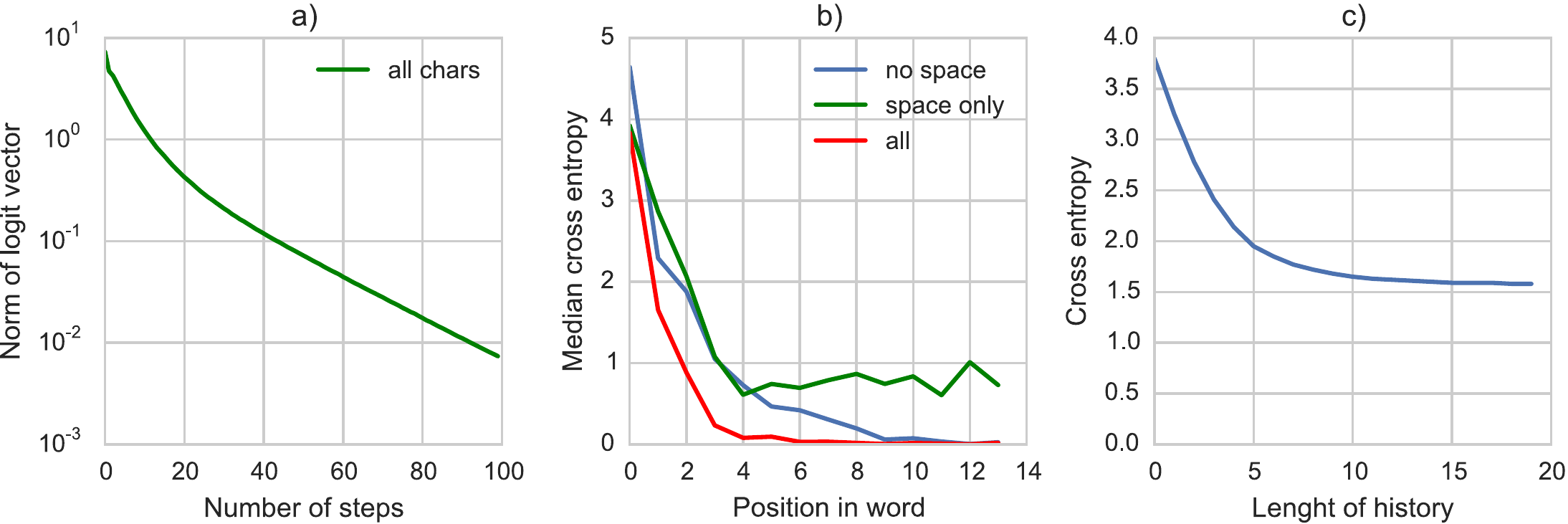}
\caption{
The time decay of the contributions from each character to prediction.
{\em a)} Average norm of $\mkappa^t_s$ across training text, $\expect{}{\norm{\mkappa^t_s}_2}$, 
plotted as a function of $t-s$, and averaged across all source characters. The norm appears to decay exponentially at two rates, a faster rate for the first ten or so characters, and then a slower rate for more long term contributions. 
{\em b)} The median cross entropy as a function of the position in the word under three different circumstances: the red line uses all of the $\mkappa^t_s$ (baseline), the green line sets all  $\mkappa^t_s$ apart from $\mkappa^t_\text{`\_'}$ to zero, while the blue line only sets $\mkappa^t_\text{`\_'}$ to zero.  The results from panel b demonstrate the disproportionately large importance of `\_' in decoding, especially at the onset of a word.
{\em c)} The cross-entropy as a function of history when artificially limiting the number of characters available for prediction. This corresponds to only considering the most recent $n$ of the $\mkappa$, where $n$ is the length of the history.
}
\label{fig:norm_decay}
\end{figure}

Using ISAN, we can investigate information timescales in the network.  For example, we investigated how quickly the contributions of $\mkappa^t_s$ decay as a function of $t-s$ on average. Figure~\ref{fig:norm_decay}a shows that this contribution decays on two different exponential timescales. We hypothesize that the first time scale corresponds to the decay within a word, while the next corresponds to the decay of information across words and sentences.  We also show the relevance of the $\mkappa_s^t$ contributions to the decoding of characters at different positions in the word (Figure~\ref{fig:norm_decay}b). For example, we observe that $\mkappa_\text{`\_'}^t$ makes important contributions to the prediction of the next character at time $t$. We show that using only the $\mkappa_\text{`\_'}^t$, the model can achieve a cross entropy of less than $1$ bit / char when the position of the character is more than 3 letters from the beginning of the word.  Finally, we link the norm-decay of $\mkappa^t_s$ to the importance of past characters for the decoding quality ( Figure~\ref{fig:norm_decay}c). By artificially limiting the number of past $\mkappa$ available for prediction we show that the prediction quality improves rapidly when extending the history from 0 to 10 characters and then saturates. This rapid improvement aligns with the range of faster decay in  Figure~\ref{fig:norm_decay}a. 

\subsection{From characters to words}
\label{sec char2word}

\begin{figure}[t]
\centering
\includegraphics[width=1\linewidth]{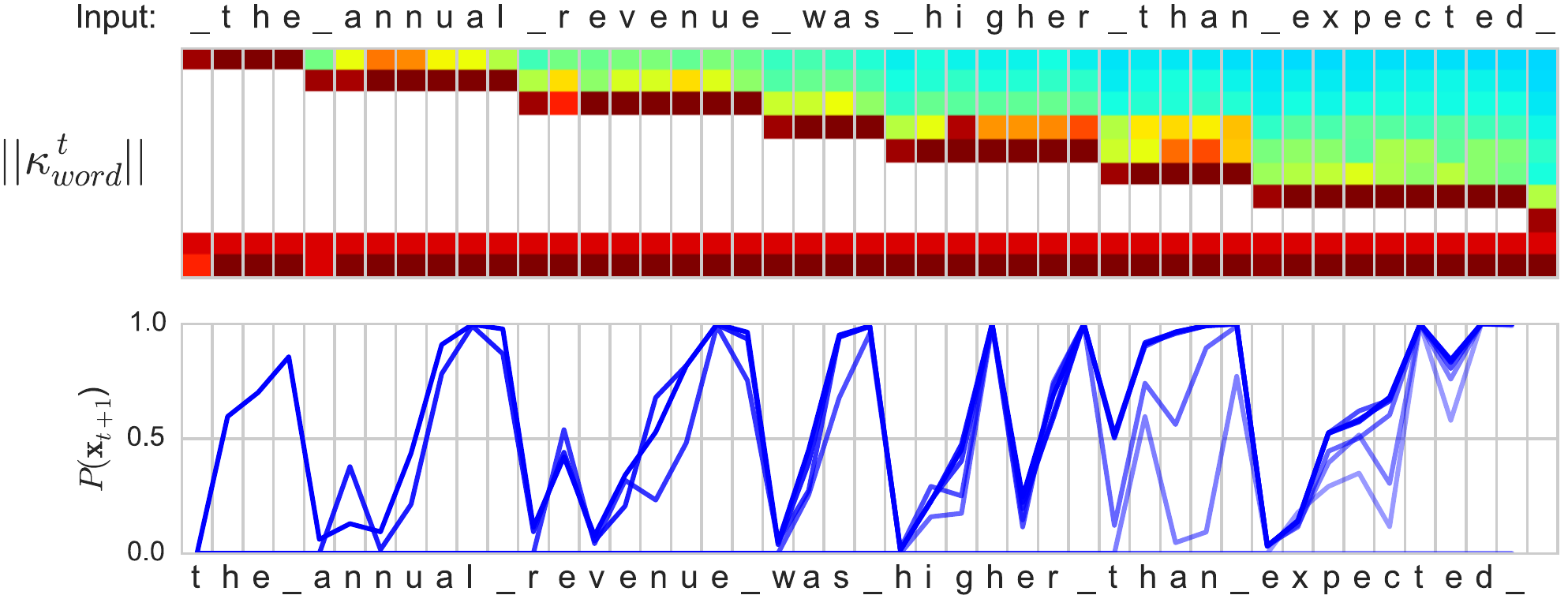}
\caption{The ISAN architecture can be used to precisely characterize the relationship between words and characters. The top panel shows how exploiting the linearity of the network's operation we can combine the $\mkappa^t_{s_1}..\mkappa^t_{s_n}$ in a word to a single contribution, $\mkappa^t_{word}$, for each word. Shown is the norm of $\mkappa^t_{word}$, a measure of the magnitude of the effect of the previous word on the selection of the current character (red corresponds to a norm of $10$, blue to $0$).  The bottom panel shows the probabilities assigned by the network to the next sequence character. Lighter lines show predictions conditioned on a decreasing number of preceding words. For example, when predicting the characters of `than' there is a large contribution from both $\mkappa^t_\text{`was'}$ and $\mkappa^t_\text{`higher'}$, as shown in the top pane. The effect on the log probabilities can be seen in the bottom panel as the model becomes less confident when excluding $\mkappa^t_\text{`was'}$ and significantly less confident when excluding both $\mkappa^t_\text{`was'}$ and $\mkappa^t_\text{`higher'}$. This word based representation clearly shows that the system leverages contextual information across multiple words. 
}
\label{fig:word_united_states}
\end{figure}

The ISAN provides a natural means of moving from character level representation to word level. Using the linearity of the hidden state dynamics we can aggregate all of the $\mkappa^t_s$ belonging to a given word and visualize them as a single contribution to the prediction of the letters in the next word. This allows us to understand how each preceding word impacts the decoding for the letters of later words. In Figure~\ref{fig:word_united_states} we show that the words `was' and `higher' make large contributions to the prediction of the characters in `than' as measured by the norm of the $\mkappa^t_\text{`\_was'}$ and $\mkappa^t_\text{`\_higher'}$.

\begin{figure}[t]
\centering
\includegraphics[width=1.0\linewidth]{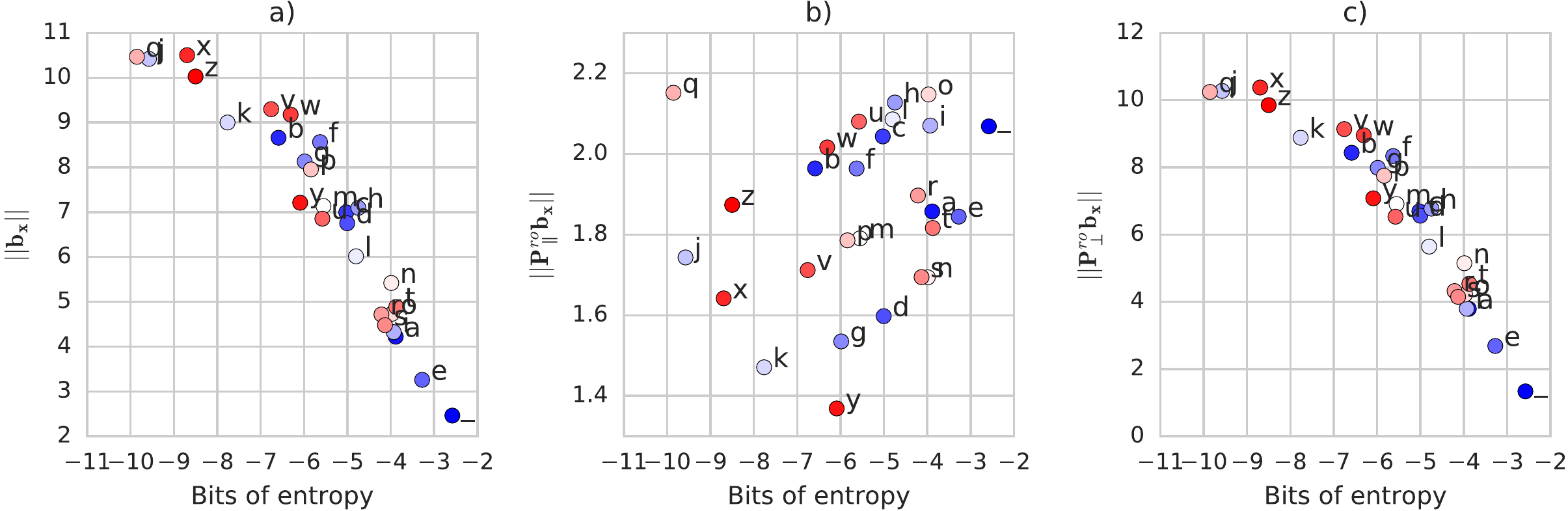}
\caption{%
By transforming the ISAN dynamics into a new basis, we can better understand the action of the input-dependent biases. 
{\em a)} We observe a strong correlation between the norms of the input dependent biases, $\mb b_{\mb x}$, and the log-probability of the unigram $\mb x$ in the training data. We can begin to understand this correlation structure using a basis transform into the `readout basis'. Breaking out the norm into its components in $\mb P^{ro}_\parallel$ and $\mb P^{ro}_\perp$ in {\em b)} and {\em c)} respectively, shows that the correlation is due to the component orthogonal to $\mb W_{ro}$.  This implies a connection between information or `surprise' and distance in the 'computational' subspace of state space.}
\label{fig:norm_broken_out}
\end{figure}

\subsection{Change of basis}\label{sec:change_of_basis}

We are free to perform a change of basis on the hidden state, and then to run the affine ISAN dynamics in that new basis.
Note that this change of basis is not possible for other RNN architectures, since the action of the nonlinearity depends on the choice of basis.

In particular we can construct a `readout basis' that explicitly divides the latent space into a subspace $\mb P^{ro}_\parallel$ spanned by the rows of the readout matrix $\mb W_{ro}$, and its orthogonal complement $\mb P^{ro}_\perp$. 
This representation explicitly divides the hidden state dynamics into a 27-dimensional `readout' subspace that is accessed by the readout matrix to make predictions, and a `computational' subspace comprising the remaining $216 - 27$ dimensions that are orthogonal to the readout matrix. 

We apply this change of basis to analyze an intriguing observation about the hidden offsets $\mb b_{\mb x}$. As shown in  Figure~\ref{fig:norm_broken_out}, the norm of the $\mb b_{\mb x}$ is strongly correlated to the log-probability of the unigram $\mb x$ in the training data. Re-expressing network parameters using the `readout basis' shows that this correlation is not related to reading out the next-step prediction.  This is because the norm of the projection of $\mb b_{\mb x}$ into $\mb P^{ro}_\perp$ remains strongly correlated with character frequency, while the projection into $\mb P^{ro}_\parallel$ shows little correlation. This indicates that the information content or 'surprise' of a letter is encoded through the norm of the component of $\mb b_{\mb x}$ in the computational space, rather than in the readout space.

\begin{figure}
\centering
\includegraphics[clip,width=1.0\linewidth]{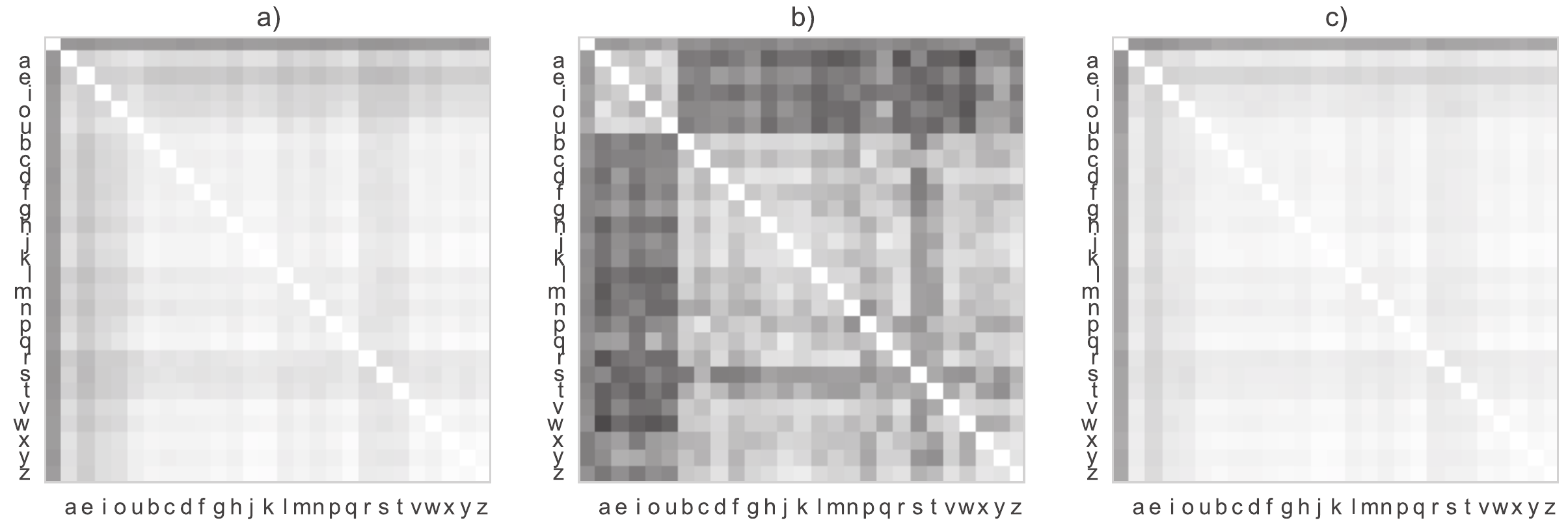}
\caption{
By transforming ISAN dynamics into a new basis, we can better interpret structure in the input-dependent biases. 
In {\em a)} we show the cosine distance between the input dependent bias vectors, split between vowels and consonants (` ' is first). In {\em b)} we show the correlation only considering the components in the subspace $\mb P^{ro}_\parallel$ spanned by the rows of the readout matrix $\mb W_{ro}$. {\em c)} shows the correlation of the components in the orthogonal complement $\mb P^{ro}_\perp$. In all plots white corresponds to 0 (aligned) and black to 2.
\jascha{add color bar}
}
\label{fig:corr_broken_out}
\end{figure}

Similarly, in Figure~\ref{fig:corr_broken_out} we illustrate that the structure in the correlations between the biases $\mb b_{\mb x}$ (across all $\mb x$) is due to their components in $\mb P^{ro}_\parallel$, while the correlation in $\mb P^{ro}_\perp$ is relatively uniform. We can clearly see two blocks of high correlations between the vowels and consonants respectively, while $\mb b_\text{`\_'}$ is uncorrelated to either.

\subsection{Comparison with $n$-gram model with back-off}

\begin{figure}[t]
\centering
\includegraphics[trim={0 0 0 0},clip,width=1.0\linewidth]{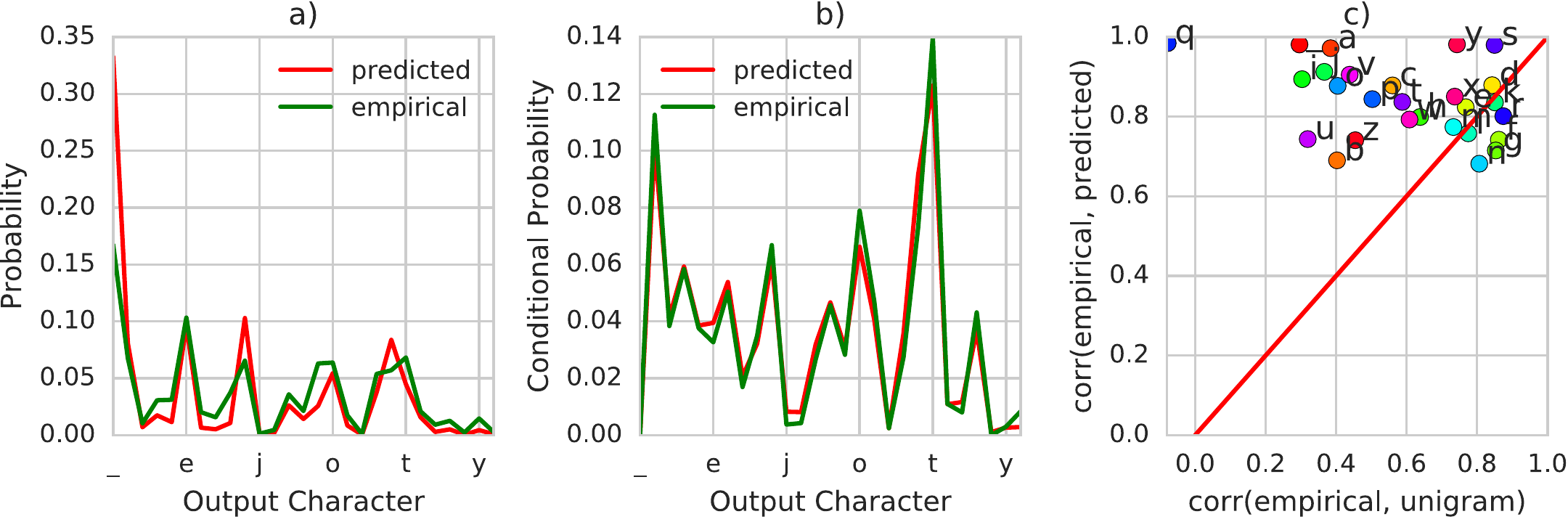}
\caption{
The predictions of ISAN for one and two characters well approximate the predictions of unigram and bigram models. 
In {\em a)} we compare softmax($\mb b_{ro}$) to the empirical unigram distribution $P(\mb x)$. In {\em b)} we compare softmax($\mb W_{ro} \mb b_\text{`\_'} + \mb b_{ro}$) with the empirical distribution $P(\mb x_{t+1} |\text{`\_'})$. In {\em c)} we show the correlation of softmax($\mb W_{ro} \mb b_{\mb x} + \mb b_{ro}$) with $P(\mb x_{t+1} | \mb x_t)$ for all 27 characters (y-axis), and compare this to the correlation between the empirical unigram probabilities $P(\mb x)$ to $P(\mb x_{t+1} | \mb x_t)$ (x-axis). The plot shows that the readout of the bias vector is a better predictor of the conditional distribution than the unigram probability. }
\label{fig:unigram_corr}
\end{figure}

We compared the computation performed by $n$-gram language models and those performed by the ISAN. An n-gram model with back-off weights expresses the conditional probability $p\pp{\mb x_t|\mb x_1... \mb x_{t-1}}$ as a sum of smoothed count ratios of $n$-grams of different lengths, with the contribution of shorter $n$-grams down-weighted by back-off weights. On the other hand, the computations performed by the ISAN start with the 
contribution of $\mb b_{ro}$ to the logits, which as shown in Figure~\ref{fig:unigram_corr}a, corresponds to the unigram log-probabilities. 
The logits are then additively updated with contributions from longer $n$-grams, represented by $\mkappa^t_{s}$. 
This additive contribution to the logits corresponds to a multiplicative modification of the emission probabilities from histories of different length. 
For long time lags, the additive correction to log-probabilities becomes small (Figure \ref{fig:opening_the_black_box}), which corresponds to multiplication by a uniform distribution. 
Despite these differences in how n-gram history is incorporated, we nevertheless observe an agreement between empirical models estimated on the training set and model predictions for unigrams and bigrams. Figure~\ref{fig:unigram_corr} shows that the bias term $\mb b_{ro}$ gives the unigram probabilities of letters, while the addition of the offset terms $\mb b_{\mb x}$ accurately predict the bigram distribution of $P\pp{\mb x_{t+1}|\mb x_t}$. Shown in panel b is an example, $P\pp{\mb x|`\_'}$, and in panel c, a summary plot for all 27 letters. 

We further explore the $n$-gram comparison by artificially limiting the length of the character history that is available to the ISAN for making predictions, as shown in Figure~\ref{fig:norm_decay}c). 

\section{Analyses of a parentheses counting task}

To show the possibility of complete interpretability of the ISAN we train a model on a parenthesis counting task. Bringing together ideas from section \ref{sec:change_of_basis} we re-express the transition dynamics in a new basis that fully reveals performed computations.

We analyze the task of counting the nesting levels of multiple parentheses types, a simplified version of a task defined in \citep{collins2016capacity}.  Briefly, a 35-unit ISAN is required to keep track of the nesting level of 2 different types of parentheses independently.  The inputs are the one-hot encoding of the different opening and closing parentheses (e.g.  `(', `)', `[', `]') as well as a noise character (`a').  The output is the one-hot encoding of the nesting level between (0-5), one set of counts for each parenthesis type (so the complete output vector is a 12 dimensional 2-hot vector). Furthermore, the target output is the nesting level {\em at the previous time step}. This artificial delay requires the model to develop a memory. One change from \citep{collins2016capacity} is that we exchange the cross-entropy error with an $L2$ error. This leads to slightly cleaner figures, but does not qualitatively change the results.

To elucidate the mechanism of ISAN's operation we first re-express the affine transitions $\mb h_{t+1} = \mb W \mb h_t +  \mb b$ by their linear equivalents $\mb h'_{t+1} = \mb W' \mb h'_t$, where ${\mb W'}  = [{\mb W} \ {\mb b}; \mb 0^T \ 1]$ and $\mb h'_t = [\mb h_t; 1]$. Next, we used linear regression to find a change of basis for which all augmented character matrices and the hidden states are sparse. To do this we construct the `readout' ($\mb P^{ro}_\parallel$) and `computational' ( $\mb P^{ro}_\perp$) subspace decomposition as discussed in Section~\ref{sec:change_of_basis}. We choose a basis for $\mb P^{ro}_\perp$ which makes the projections of the hidden states into this computational subspace 2-hot vectors. With this subspace decomposition, the hidden states and character matrices have the form

\begin{equation} 
\WW_x' =  \begin{bmatrix}
\WW_x^{rr} \; \WW_x^{rc} \; \bb_x^r \\
\WW_x^{cr} \; \WW_x^{cc} \; \bb_x^c \\
\mathbf{0}^T \; \phantom{a}\mathbf{0}^T \; \phantom{a}1
\end{bmatrix}
\quad \hh_t' =  \begin{bmatrix}
\hh_t^{r} \\
\hh_t^{c} \\
1
\end{bmatrix}
\end{equation}

and the update equation can be written as

\begin{equation}
\quad \hh_{t+1}' = \WW_x' \hh_t' = \begin{bmatrix}
\WW_x^{rr} \hh_t^r + \WW_x^{rc} \hh_t^c + \bb_x^r \\
\WW_x^{cr} \hh_t^r + \WW_x^{cc} \hh_t^c + \bb_x^c \\
1
\end{bmatrix}.
\end{equation}

Here $\mb h_t^r$ and $\mb h_t^c$ denote the readout and computational portions of $\mb h_t$, and $\WW_x^{rr}, \WW_x^{cr}, \WW_x^{rc}, \WW_x^{cc}$ denote the readout to readout, readout to computation, computation to readout, and computation to computation blocks of the character matrix for character $x$, respectively. 

\begin{figure}[t]
\centering
\includegraphics[trim={.5cm .5cm .5cm .5cm},clip,width=1.0\linewidth]{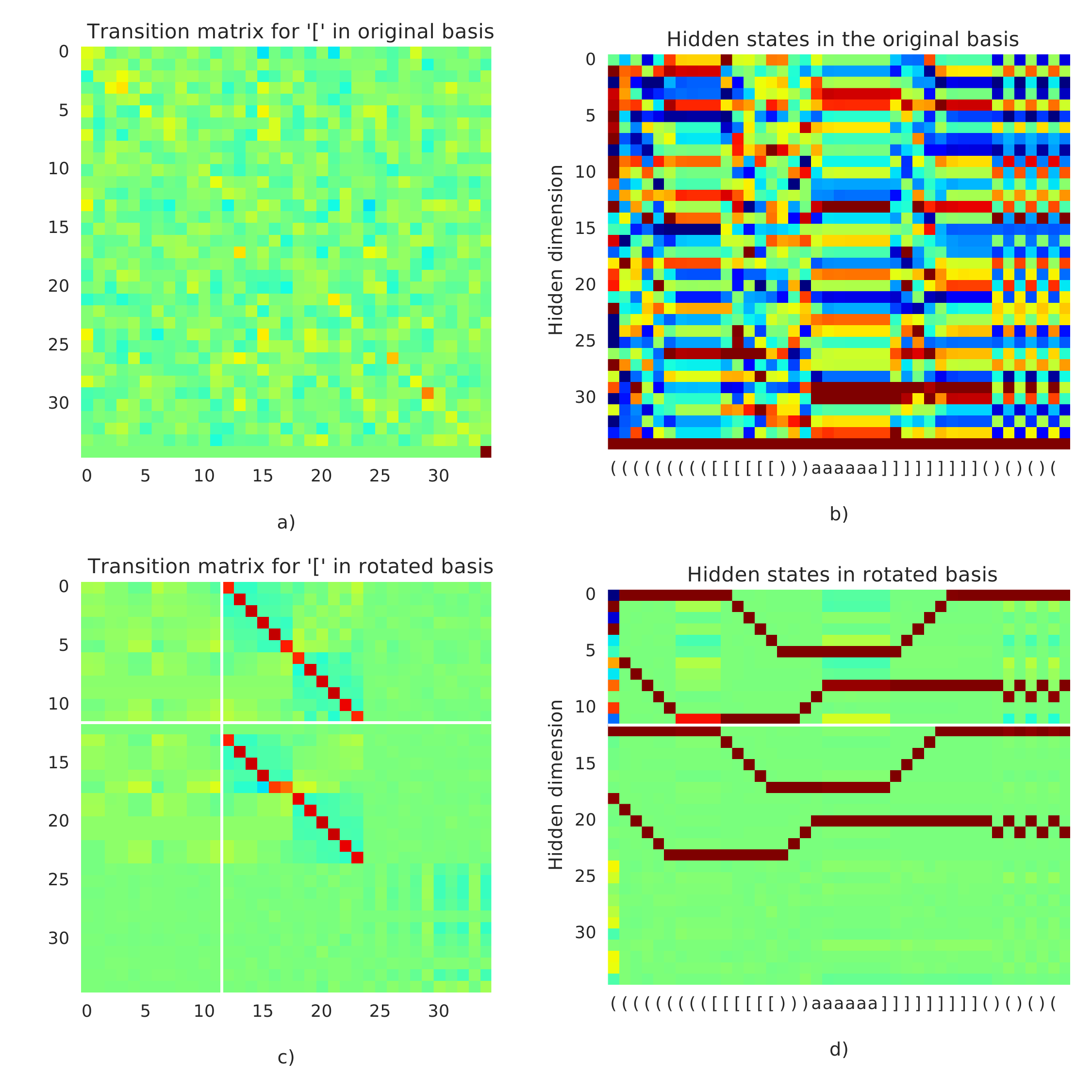} 
\caption{%
A visualization of the dynamics of an ISAN for the two parentheses counting task with 1 time lag (count either '()' or '[]' nesting levels with a one-step readout delay). 
In {\em a)} the weight matrix for `[' is shown in the original basis.  In {\em c)} it is shown transformed to highlight the delay-line dynamics. The activations of the hidden units are shown {\em b)} in the original basis, and {\em d)} rotated to the same basis as in c), to highlight the delay-line dynamics in a more intelligible way. The white line delineates the transition matrix elements and hidden state dimensions that directly contribute to the output. All matrices for parentheses types appear similarly, with closing parentheses, e.g. `]', changing the direction of the delay line.}
\label{fig:lag_task}
\end{figure}

In Figure~\ref{fig:lag_task}d we show the hidden states in the rotated basis as a sequence of column vectors. The 35 dimensional hidden states are all 4-hot. We can treat them as a concatenation of a readout $\mb h_t^r$ and a computation $\mb h_t^c$ part. The 12-dimensional readout $\mb h_t^r$ corresponds to network's output at time step $t$ and encodes the counts from time step $t-1$ as a 2-hot vector (one count per parenthesis type). The computational space $\mb h_t^c$ is $35 - 12 = 23$ dimensional, and encodes the current counts as another 2-hot vector. Note that in this basis the ISAN effectively uses only 24 dimensions and the remaining 11 dimensions have no noticeable effect on the computation.  In Figure~\ref{fig:lag_task}c we show $\WW_{[}'$ in the rotated basis. We see from the leftmost 12 columns that $\WW_{[}^{rr}$ and $\WW_{[}^{cr}$ are both nearly 0. This means that $\mb h_t^r$ has no influence on $\mb h_{t+1}$. Furthermore, the computation to readout block, $\WW_{[}^{rc}$, is identity on the first 12 dimensions, effectively implementing the lagging output $\mb h_{t}^r = \mb h_{t-1}^c$. The current counts are implemented as delay lines and identity sub-matrices in $\WW_{[}^{cc}$, which respectively has the effect of incrementing the count of `[' by one, saturating at 5, and leaving the count of '()' parentheses fixed.  The matrices $\WW_{]}, \WW_{(}, \WW_{)}$ behave analogously.  It is clear that this solution is general, in that retraining for increased numbers of parentheses types or an increased counting maximum, would have the analogous solution.

\section{Discussion}

In this paper we motivated an input-switched affine recurrent network for the purpose of interpretability.  We showed that a switched affine architecture achieves the same performance as LSTMs on the Text8 dataset for the same number of maximum parameters, and reasonable performance on the BWB.  We performed a series of analyses, demonstrating the ability to understand how inputs at one point in the input sequence affect the outputs later in the output sequence.  We showed further in the multiple parentheses counting task that the ISAN dynamics can be completely reverse engineered.  In summary, this work provides evidence that the ISAN is able to express complex dynamical systems, yet its operation can in principle be fully understood, a prospect that remains out of reach for many popular recurrent architectures.

\subsection{Computational benefits}\label{sec:comp}

Switched affine networks hold the potential to be massively more computationally and memory efficient for text processing than other recurrent architectures. 
First, input-dependent affine transitions reduce the number of parameters used at every step.
For $K$ possible inputs and $N$ parameters, the computational cost per update step is $O\pp{\frac{N}{K}}$, a factor of $K$ speedup over non-switched architectures.
Similarly, the number of hidden units is $O\pp{\sqrt{\frac{N}{K}}}$, a factor of $K^\frac{1}{2}$ memory improvement for storage of the latent state.

Furthermore, the ISAN is unique in its ability to pre-compute affine transformations corresponding to input strings. This is possible because the composition of affine transformations is also an affine transformation. 
This property is used in Section \ref{sec char2word} to evaluate the linear contributions of words, rather than characters. This means that the hidden state update corresponding to an entire input sequence can be computed with identical cost to the update for a single character (plus the dictionary look-up cost for the composed transformation). ISAN can therefore achieve very large speedups on input processing, at the cost of increased memory use, by accumulating large look-up tables of the $\mb W_{\mb x}$ and $\mb b_{\mb x}$ corresponding to common input sequences.  Of course, practical implementations will have to incorporate complexities of memory management, batching, etc.

\subsection{Future work}

There are some obvious future directions to this work.  Currently, we define switching behavior using an input set with finite and manageable cardinality.  Studying word-level language models with enormous vocabularies may require some additional logic to scale.  Another idea is to build a language model that switches on bigrams or trigrams, rather than characters or words, targeting an intermediate number of affine transformations.  Adapting this model to continuous-valued inputs is another important direction.  One approach is to use a tensor factorization similar to that employed by the MRNN \citep{sutskever2014sequence} or defining weights via additional networks, as in HyperNetworks \citep{ha2016hypernetworks}. Finally, we expect that automated methods for changing bases to enable sparse representations in the hidden state and dynamics matrices will be a particularly fruitful direction to pursue.

%
\section*{Acknowledgements}
We would like to thank Jasmine Collins for her help and advice, and Quoc Le, David Ha and Mohammad Norouzi for helpful discussions. We would also like to thank Herbert Jaeger for insightful discussions regarding the Observable-Operator-Model.

\bibliography{ISAN}

\begin{thebibliography}{34}
\providecommand{\natexlab}[1]{#1}
\providecommand{\url}[1]{\texttt{#1}}
\expandafter\ifx\csname urlstyle\endcsname\relax
  \providecommand{\doi}[1]{doi: #1}\else
  \providecommand{\doi}{doi: \begingroup \urlstyle{rm}\Url}\fi

\bibitem[Alain \& Bengio(2016)Alain and Bengio]{alain2016understanding}
Alain, Guillaume and Bengio, Yoshua.
\newblock Understanding intermediate layers using linear classifier probes.
\newblock \emph{arXiv preprint arXiv:1610.01644}, 2016.

\bibitem[Andrew(2013)]{andrew2013introduction}
Andrew, Alex~M.
\newblock An introduction to support vector machines and other kernel-based
  learning methods.
\newblock \emph{Kybernetes}, 2013.

\bibitem[Belanger \& Kakade(2015)Belanger and Kakade]{belanger2015linear}
Belanger, David and Kakade, Sham.
\newblock A linear dynamical system model for text.
\newblock In \emph{Proceedings of the 32nd International Conference on Machine
  Learning (ICML-15)}, pp.\  833--842, 2015.

\bibitem[Berk et~al.(2017)Berk, Heidari, Jabbari, Kearns, and
  Roth]{berk2017fairness}
Berk, Richard, Heidari, Hoda, Jabbari, Shahin, Kearns, Michael, and Roth,
  Aaron.
\newblock Fairness in criminal justice risk assessments: The state of the art.
\newblock \emph{arXiv preprint arXiv:1703.09207}, 2017.

\bibitem[Bojarski et~al.(2016)Bojarski, Del~Testa, Dworakowski, Firner, Flepp,
  Goyal, Jackel, Monfort, Muller, Zhang, et~al.]{bojarski2016end}
Bojarski, Mariusz, Del~Testa, Davide, Dworakowski, Daniel, Firner, Bernhard,
  Flepp, Beat, Goyal, Prasoon, Jackel, Lawrence~D, Monfort, Mathew, Muller,
  Urs, Zhang, Jiakai, et~al.
\newblock End to end learning for self-driving cars.
\newblock \emph{arXiv preprint arXiv:1604.07316}, 2016.

\bibitem[{Chelba} et~al.(2013){Chelba}, {Mikolov}, {Schuster}, {Ge}, {Brants},
  {Koehn}, and {Robinson}]{2013arXiv1312.3005C}
{Chelba}, C., {Mikolov}, T., {Schuster}, M., {Ge}, Q., {Brants}, T., {Koehn},
  P., and {Robinson}, T.
\newblock {One Billion Word Benchmark for Measuring Progress in Statistical
  Language Modeling}.
\newblock \emph{ArXiv e-prints}, December 2013.

\bibitem[Ching et~al.(2017)Ching, Himmelstein, Beaulieu-Jones, Kalinin, Do,
  Way, Ferrero, Agapow, Xie, Rosen, et~al.]{ching2017opportunities}
Ching, Travers, Himmelstein, Daniel~S, Beaulieu-Jones, Brett~K, Kalinin,
  Alexandr~A, Do, Brian~T, Way, Gregory~P, Ferrero, Enrico, Agapow,
  Paul-Michael, Xie, Wei, Rosen, Gail~L, et~al.
\newblock Opportunities and obstacles for deep learning in biology and
  medicine.
\newblock \emph{bioRxiv}, pp.\  142760, 2017.

\bibitem[Collins et~al.(2016)Collins, Sohl-Dickstein, and
  Sussillo]{collins2016capacity}
Collins, Jasmine, Sohl-Dickstein, Jascha, and Sussillo, David.
\newblock Capacity and trainability in recurrent neural networks.
\newblock \emph{ICLR 2017 submission}, 2016.

\bibitem[{Council of European Union}(2016)]{eureg_2016_679}
{Council of European Union}.
\newblock {General Data Protection Regulation, Article 22 (Regulation (EU)
  2016/679)}, 2016.
\newblock URL \url{http://www.privacy-regulation.eu/en/22.htm}.

\bibitem[Deo(2015)]{deo2015machine}
Deo, Rahul~C.
\newblock Machine learning in medicine.
\newblock \emph{Circulation}, 132\penalty0 (20):\penalty0 1920--1930, 2015.

\bibitem[Freedman(2009)]{freedman2009statistical}
Freedman, David~A.
\newblock \emph{Statistical models: theory and practice}.
\newblock cambridge university press, 2009.

\bibitem[Gulshan et~al.(2016)Gulshan, Peng, Coram, Stumpe, Wu, Narayanaswamy,
  Venugopalan, Widner, Madams, Cuadros, et~al.]{gulshan2016development}
Gulshan, Varun, Peng, Lily, Coram, Marc, Stumpe, Martin~C, Wu, Derek,
  Narayanaswamy, Arunachalam, Venugopalan, Subhashini, Widner, Kasumi, Madams,
  Tom, Cuadros, Jorge, et~al.
\newblock Development and validation of a deep learning algorithm for detection
  of diabetic retinopathy in retinal fundus photographs.
\newblock \emph{JAMA}, 316\penalty0 (22):\penalty0 2402--2410, 2016.

\bibitem[Ha et~al.(2016)Ha, Dai, and Le]{ha2016hypernetworks}
Ha, David, Dai, Andrew, and Le, Quoc~V.
\newblock Hypernetworks.
\newblock \emph{arXiv preprint arXiv:1609.09106}, 2016.

\bibitem[Hochreiter \& Schmidhuber(1997)Hochreiter and
  Schmidhuber]{hochreiter1997long}
Hochreiter, Sepp and Schmidhuber, J{\"u}rgen.
\newblock Long short-term memory.
\newblock \emph{Neural computation}, 9\penalty0 (8):\penalty0 1735--1780, 1997.

\bibitem[Hwang \& Sung(2016)Hwang and Sung]{hwang16character}
Hwang, Kyuyeon and Sung, Wonyong.
\newblock Character-level language modeling with hierarchical recurrent neural
  networks.
\newblock \emph{CoRR}, abs/1609.03777, 2016.
\newblock URL \url{http://arxiv.org/abs/1609.03777}.

\bibitem[Jaeger(2000)]{jaeger2000observable}
Jaeger, Herbert.
\newblock Observable operator models for discrete stochastic time series.
\newblock \emph{Neural Computation}, 12\penalty0 (6):\penalty0 1371--1398,
  2000.

\bibitem[J{\'{o}}zefowicz et~al.(2016)J{\'{o}}zefowicz, Vinyals, Schuster,
  Shazeer, and Wu]{jozefowicz2016exploring}
J{\'{o}}zefowicz, Rafal, Vinyals, Oriol, Schuster, Mike, Shazeer, Noam, and Wu,
  Yonghui.
\newblock Exploring the limits of language modeling.
\newblock \emph{CoRR}, abs/1602.02410, 2016.
\newblock URL \url{http://arxiv.org/abs/1602.02410}.

\bibitem[Karpathy et~al.(2015)Karpathy, Johnson, and
  Li]{karpathy2015visualizing}
Karpathy, Andrej, Johnson, Justin, and Li, Fei-Fei.
\newblock Visualizing and understanding recurrent networks.
\newblock \emph{arXiv preprint arXiv:1506.02078}, 2015.

\bibitem[Katz et~al.(2017)Katz, Barrett, Dill, Julian, and
  Kochenderfer]{katz2017reluplex}
Katz, Guy, Barrett, Clark, Dill, David, Julian, Kyle, and Kochenderfer, Mykel.
\newblock Reluplex: An efficient smt solver for verifying deep neural networks.
\newblock \emph{arXiv preprint arXiv:1702.01135}, 2017.

\bibitem[Le et~al.(2012)Le, Ranzato, Monga, Devin, Chen, Corrado, Dean, and
  Ng]{le2012building}
Le, Quoc~V, Ranzato, Marc~A., Monga, Rajat, Devin, Matthieu, Chen, Kai,
  Corrado, Greg~S., Dean, J, and Ng, Andrew~Y.
\newblock Building high-level features using large scale unsupervised learning.
\newblock In \emph{International Conference on Machine Learning}, 2012.

\bibitem[Linderman et~al.(2016)Linderman, Miller, Adams, Blei, Paninski, and
  Johnson]{linderman2016recurrent}
Linderman, Scott~W, Miller, Andrew~C, Adams, Ryan~P, Blei, David~M, Paninski,
  Liam, and Johnson, Matthew~J.
\newblock Recurrent switching linear dynamical systems.
\newblock \emph{arXiv preprint arXiv:1610.08466}, 2016.

\bibitem[Mahoney(2011)]{text8}
Mahoney, Matt.
\newblock Large text compression benchmark: About the test data, 2011.
\newblock URL \url{http://mattmahoney.net/dc/textdata}.
\newblock [Online; accessed 15-November-2016].

\bibitem[Martens \& Sutskever(2011)Martens and Sutskever]{martens2011learning}
Martens, James and Sutskever, Ilya.
\newblock Learning recurrent neural networks with hessian-free optimization.
\newblock In \emph{Proceedings of the 28th International Conference on Machine
  Learning (ICML-11)}, pp.\  1033--1040, 2011.

\bibitem[Mikolov et~al.(2012)Mikolov, Sutskever, Deoras, Le, and
  Kombrink]{mikolov2012subword}
Mikolov, Tom{\'a}{\v{s}}, Sutskever, Ilya, Deoras, Anoop, Le, Hai-Son, and
  Kombrink, Stefan.
\newblock Subword language modeling with neural networks.
\newblock \emph{preprint}, 2012.

\bibitem[Mordvintsev et~al.(2015)Mordvintsev, Olah, and
  Tyka]{mordvintsev2015inceptionism}
Mordvintsev, Alexander, Olah, Christopher, and Tyka, Mike.
\newblock Inceptionism: Going deeper into neural networks.
\newblock \emph{Google Research Blog. Retrieved June}, 20:\penalty0 14, 2015.

\bibitem[Murdoch \& Szlam(2017)Murdoch and Szlam]{murdoch2017automatic}
Murdoch, W.~James and Szlam, Arthur.
\newblock Automatic rule extraction from long short term memory networks.
\newblock In \emph{ICLR}, 2017.

\bibitem[Quinlan(1987)]{quinlan1987simplifying}
Quinlan, J.~Ross.
\newblock Simplifying decision trees.
\newblock \emph{International journal of man-machine studies}, 27\penalty0
  (3):\penalty0 221--234, 1987.

\bibitem[Scarborough \& Somers(2006)Scarborough and
  Somers]{scarborough2006neural}
Scarborough, David and Somers, Mark~John.
\newblock \emph{Neural networks in organizational research: Applying pattern
  recognition to the analysis of organizational behavior.}
\newblock American Psychological Association, 2006.

\bibitem[Siano et~al.(2012)Siano, Cecati, Yu, and Kolbusz]{siano2012real}
Siano, Pierluigi, Cecati, Carlo, Yu, Hao, and Kolbusz, Janusz.
\newblock Real time operation of smart grids via fcn networks and optimal power
  flow.
\newblock \emph{IEEE Transactions on Industrial Informatics}, 8\penalty0
  (4):\penalty0 944--952, 2012.

\bibitem[Sussillo \& Barak(2013)Sussillo and Barak]{sussillo2013opening}
Sussillo, David and Barak, Omri.
\newblock Opening the black box: low-dimensional dynamics in high-dimensional
  recurrent neural networks.
\newblock \emph{Neural computation}, 25\penalty0 (3):\penalty0 626--649, 2013.

\bibitem[Sutskever et~al.(2011)Sutskever, Martens, and
  Hinton]{sutskever2011generating}
Sutskever, Ilya, Martens, James, and Hinton, Geoffrey~E.
\newblock Generating text with recurrent neural networks.
\newblock In \emph{Proceedings of the 28th International Conference on Machine
  Learning (ICML-11)}, pp.\  1017--1024, 2011.

\bibitem[Sutskever et~al.(2014)Sutskever, Vinyals, and
  Le]{sutskever2014sequence}
Sutskever, Ilya, Vinyals, Oriol, and Le, Quoc~V.
\newblock Sequence to sequence learning with neural networks.
\newblock In \emph{Advances in neural information processing systems}, pp.\
  3104--3112, 2014.

\bibitem[Tashea(2017)]{tashea2017}
Tashea, Jason.
\newblock Courts are using ai to sentence criminals. that must stop now.
\newblock \emph{WIRED magazine}, 2017.

\bibitem[Zeiler et~al.(2010)Zeiler, Krishnan, Taylor, and
  Fergus]{zeiler2010deconvolutional}
Zeiler, Matthew~D, Krishnan, Dilip, Taylor, Graham~W, and Fergus, Rob.
\newblock Deconvolutional networks.
\newblock In \emph{Computer Vision and Pattern Recognition (CVPR), 2010 IEEE
  Conference on}, pp.\  2528--2535. IEEE, 2010.

\end{thebibliography}
\bibliographystyle{icml2017}

\end{document}